\documentclass{article}

 \usepackage[sglblindworkshop, final]{latinx_2025}

\usepackage[utf8]{inputenc} 
\usepackage[T1]{fontenc}    
\usepackage{hyperref}       
\usepackage{url}            
\usepackage{booktabs}       
\usepackage{amsfonts}       
\usepackage{nicefrac}       
\usepackage{microtype}      
\usepackage{xcolor}         
\usepackage{multirow}
\usepackage{graphicx}
\usepackage[table]{xcolor}
\usepackage{float}
\usepackage{listings}

\title{Do Large Language Models Understand Data Visualization Rules?}

\usepackage{dsfont}
\usepackage{overpic}
\usepackage{contour}
\contourlength{1pt}
\usepackage{xspace}
\usepackage{sidecap}
\usepackage{rotating}
\usepackage[normalem]{ulem}
\usepackage{amsmath}
\usepackage{xcolor,colortbl}
\usepackage{tabularx}
\usepackage{soul}
\usepackage{url}
\usepackage{amsfonts}
\usepackage{comment}
\usepackage{booktabs}
\usepackage{etoolbox}

\newlength{\indentlaenge}
\author{
    \textbf{Martín Sinnona}$^{1,3}$ \hspace{0.1cm} 
    \textbf{Valentín Bonás}$^{1}$ \hspace{0.1cm}
    \textbf{Emmanuel Iarussi}$^{1,2}$ \hspace{0.1cm}
    \textbf{Viviana Siless}$^{1}$ \hspace{0.1cm} \\
    Universidad Torcuato Di Tella, Buenos Aires, Argentina $^{1}$\\
    Consejo Nacional de Investigaciones Científicas y Técnicas, Buenos Aires, Argentina $^{2}$ \\
    Universidad de Buenos Aires, Buenos Aires, Argentina $^{3}$ \\
    \texttt{msinnona@dc.uba.ar}
}

\begin{document}

\maketitle

\begin{abstract}

Data visualization rules—derived from decades of research in design and perception—ensure trustworthy chart communication. While prior work has shown that large language models (LLMs) can generate charts or flag misleading figures, it remains unclear whether they can reason about and enforce visualization rules directly. Constraint-based systems such as Draco encode these rules as logical constraints for precise automated checks, but maintaining symbolic encodings requires expert effort, motivating the use of LLMs as flexible rule validators.
In this paper, we present the first systematic evaluation of LLMs against visualization rules using hard-verification ground truth derived from Answer Set Programming (ASP). We translated a subset of Draco’s constraints into natural-language statements and generated a controlled dataset of ~2,000 Vega-Lite specifications annotated with explicit rule violations. LLMs were evaluated on both accuracy in detecting violations and prompt adherence, which measures whether outputs follow the required structured format.
Results show that frontier models achieve high adherence (Gemma 3 4B / 27B: 100\%, GPT-oss 20B: 98\%) and reliably detect common violations (F1 up to 0.82), yet performance drops for subtler perceptual rules (F1 < 0.15 for some categories) and for outputs generated from technical ASP formulations. Translating constraints into natural language improved performance by up to 150\% for smaller models. These findings demonstrate the potential of LLMs as flexible, language-driven validators while highlighting their current limitations compared to symbolic solvers.

\end{abstract}

\section{Introduction}

Clear and effective visualization is essential for trustworthy data communication. 
Decades of research in human perception, graphical integrity, and design best practices have been distilled into actionable rules that help ensure charts convey information accurately and without distortion \cite{cleveland1984graphical,huff1954lie,munzner2014visualization}. 
Authoring tools and validation frameworks have sought to operationalize these rules, enabling users to avoid misleading choices and improve chart clarity.

Constraint-based systems such as Draco~\cite{moritz2018formalizing} formalize these guidelines as logical constraints, enabling automatic checks over chart specifications. 
Similarly, linter frameworks like \emph{VizLinter}~\cite{chen2021vizlinter} carry these ideas into production by flagging (and sometimes repairing) violations in Vega-Lite specs. 
These approaches demonstrate that rules can be encoded precisely and applied automatically. 
However, encoding, extending, and maintaining such rule sets requires specialized expertise and tooling (e.g., ASP with Clingo~\cite{clingo}), limiting their scalability and flexibility.

Recent work hints that large (vision-)language models may offer an alternative. 
Lo et al.~\cite{lo2024good} show that multimodal LLMs can detect a wide range of misleading chart patterns directly from images when carefully prompted, expanding from five to twenty-one issue categories and surfacing the need for shared benchmarks and clearer evaluation protocols. 
Complementing this detection focus, Hong et al.~\cite{hong2025llms} evaluate whether LLMs possess \emph{visualization literacy}—the ability to read, interpret, and reason about visual representations—using a modified Visualization Literacy Assessment Test (VLAT). 
They find that while models like GPT-4 and Gemini can sometimes answer visualization questions correctly, their performance lags behind human baselines and often relies on prior knowledge rather than information present in the chart.

While promising, these efforts stop short of assessing whether models can reason about the design rules that govern visualization practice. 
Lo et al.~\cite{lo2024good} evaluate image-level detection of misleading patterns, while Hong et al.~\cite{hong2025llms} probe visualization literacy through question answering. 
Both perspectives highlight important capabilities, but neither tests whether models can enforce established rules in chart specifications themselves. 
For example, a model might correctly answer a comprehension question in an image, yet still allow a data visualization specification that encodes ordered data with color hue—an explicit rule violation \cite{cleveland1984graphical}.
Crucially, no existing benchmark ties such violations directly to solver-verified ground truth in visualization specifications, leaving open whether LLMs actually understand and apply visualization rules.

This paper asks: \emph{Do Large Language Models understand data-visualization rules?} 
We study whether LLMs (text-only and multimodal) can read a chart specification, reason about established rules derived from perception and design research, and accurately identify rule violations. 
To this end, we translate a subset of Draco’s ASP constraints into natural-language rules, generate a controlled dataset of 2{,}000 Vega-Lite specifications annotated with explicit violations, and evaluate multiple models for both correctness and adherence. 
This design directly targets explicit rule compliance, filling the gap between solver-based pipelines and image-only evaluations.

Our study is, to our knowledge, the first evaluation of LLMs against \emph{hard-verification rules} originally encoded as ASP constraints. 
This guarantees that every violation in our dataset has a precise, solver-verified ground truth. 
We introduce metrics for both \emph{accuracy} (did the model detect the correct rule violation?) and \emph{prompt adherence} (did the model follow the requested evaluation format?), enabling fine-grained assessment across models and prompts. Results show that frontier LLMs perform well at detecting common rule violations (F1 up to 0.82) but remain inconsistent across categories and often struggle with subtler perceptual constraints (F1 < 0.15). Prompt adherence is also model-dependent: Gemma 3 4B and 27B achieved 100\%, GPT-oss 20B reached 98\%, while Llama variants scored lower (0.65–0.87). These findings highlight the potential of LLMs as flexible, language-driven validators of visualization rules, while emphasizing their current limitations compared to symbolic solvers.

\section{Related Work}

The rapid emergence of large (vision-)language models has sparked interest in whether they can reason about visualization quality. 
Lo et al.~\cite{lo2024good} showed that multimodal LLMs can detect a wide range of misleading chart patterns across 21 categories, highlighting both the promise of LLMs for chart critique and the need for standardized benchmarks. 
Alexander et al.~\cite{alexander2024can} extended this line of inquiry by testing GPT-4 variants on real-world visualization pairs, demonstrating that accuracy improves when prompts include explicit definitions and examples. 
Hong et al.~\cite{chen2021vizlinter} took a different approach, evaluating models’ \emph{visualization literacy} using a modified VLAT instrument. 
They found that while models can sometimes answer comprehension questions correctly, their performance lags behind humans and often relies on prior knowledge rather than chart-based reasoning. 
Together, these studies suggest that LLMs exhibit partial capabilities in detecting errors or answering questions about charts, but leave open whether they can enforce the design rules that underlie effective visualization.

\paragraph{Rule-based systems as ground truth.}  
Before the rise of LLMs, visualization researchers formalized design knowledge into constraint-based systems. 
Moritz et al.~\cite{moritz2018formalizing} introduced Draco, which encodes guidelines as logical constraints and checks chart specifications for violations. 
Building on this, Chen et al.~\cite{chen2021vizlinter} developed \emph{VizLinter}, a linter-fixer pipeline that can automatically flag and repair problems in Vega-Lite specifications. 
These systems ensure precise rule checking, but require specialized tooling: constraints are implemented in ASP~\cite{vladimir2008answer} and evaluated using solvers such as Clingo~\cite{clingo}. 
As a result, extending Draco with new findings from perception studies is labor-intensive, since each rule must be manually implemented and debugged. 
Therefore, while effective as verification engines, these systems are not very flexible.

\paragraph{Interactive support with LLMs.}  
Beyond detection, LLMs are increasingly studied as interactive visualization assistants. 
Shin et al.~\cite{shin2025visualizationary} embedded LLM feedback directly into the design process, providing guideline-based prompts that suggest actionable improvements. 
Shen et al.~\cite{shen2024ask} investigated troubleshooting workflows, comparing human- and AI-assisted support. 
They found that while humans offer nuanced, context-sensitive guidance, LLMs provide faster feedback but often lack contextual grounding. 
This body of work highlights the potential of LLMs not just for critique, but also as collaborators in visualization design—though alignment with established perceptual rules remains a challenge.

\section{Method}
\begin{figure}[t!]
    \centering
    \includegraphics[width=0.61\textheight]{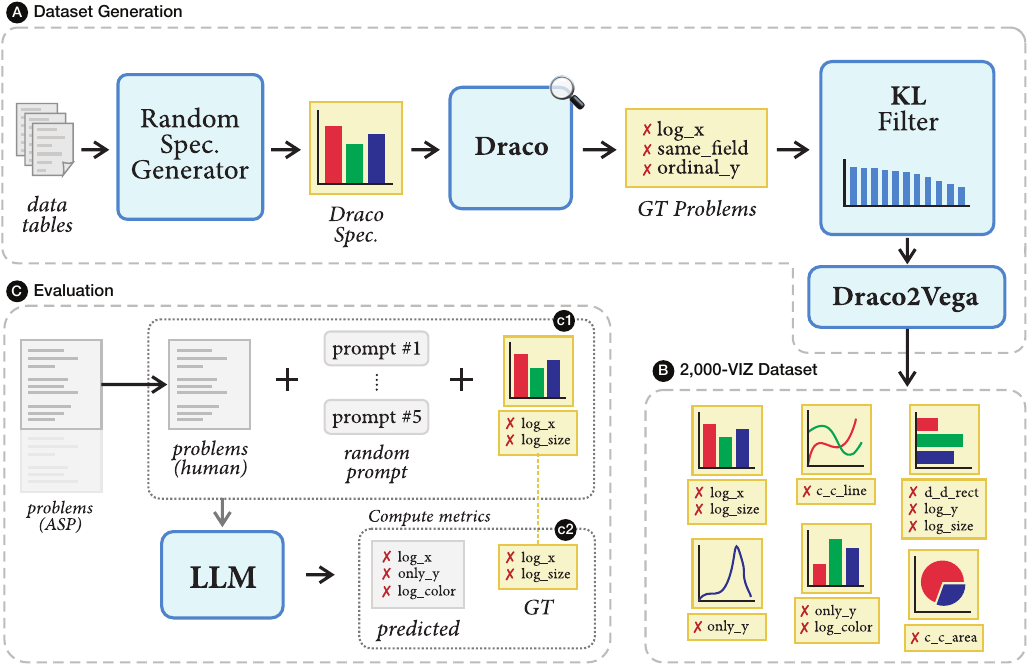}
    \caption{\textbf{Overview of the evaluation framework.} 
(\textbf{A}) We generated random chart specifications and use Draco's solver to identify ground-truth visualization problems. Apply a Kullback--Leibler (KL) divergence filter to ensure a balanced distribution of problem types. Then, we converted the accepted specifications into Vega-Lite format, yielding a dataset of 2,000 annotated instances (\textbf{B}).
(\textbf{C.1}) Subsequently, we evaluated LLMs by prompting each specification with a randomly sampled instruction prompt from five variants, describing the desired output format, and the list of possible problems. (\textbf{C.2}) Finally, we compared model predictions against the ground truth to compute accuracy and prompt adherence metrics.}
    \label{fig:overview}
\end{figure}

We evaluate the capacity of large language models (LLMs) to detect visualization design problems in Vega-Lite specifications, using Draco as the source of ground truth. Draco encodes visualization guidelines as handcrafted ASP-based constraints and preferences drawn from theoretical and empirical studies. We select a subset of these constraints—henceforth referred to as \emph{problems}—restricted to cases identifiable from chart images and descriptions alone, ensuring the focus remains on perceptually observable design rules. To facilitate evaluation, we translated the ASP constraints into human-readable rules, enabling comparison under both formal and natural language formulations.

Figure~\ref{fig:overview} outlines our evaluation framework. In stage one (Panel \textbf{A} and \textbf{B}), we depict candidate chart generation and problem identification with Draco. To balance the distribution of problem types, we apply a Kullback–Leibler divergence filter, then convert the accepted charts into Vega-Lite format, producing a dataset of 2,000 annotated instances.

In stage two (Panel \textbf{C}), we evaluate LLMs by prompting them with each Vega-Lite specification, paired with one of five randomly sampled instruction variants. Their predictions are then compared to the Draco annotations to compute accuracy and prompt-adherence metrics. The following subsections describe these stages in detail, beginning with dataset generation.

\subsection{Dataset Generation}
To our knowledge, no large-scale curated datasets of charts with labeled visualization design problems currently exist. To fill this gap, we synthetically generated 2,000 chart pairs in Draco and Vega-Lite, each annotated with the specific Draco rules they violate.

We began with 20 different datasets of data tables, comprising data from different domains, collected from Kaggle~\footnote{\url{https://www.kaggle.com/}}, and a base chart specification written in Draco grammar, for which we randomly assigned parameters such as \emph{mark type, encodings,} and \emph{variables}. Rather than exhaustively iterating over the entire parameter space, we randomly sampled configurations that were more likely to produce diverse problems, aiming to cover the entire space we were interested in. 
Each specification was then annotated with ground-truth problems identified through Draco’s automatic detection mechanism, based on Clingo solving.

\begin{figure}[t]
    \centering
    \includegraphics[width=1\linewidth]{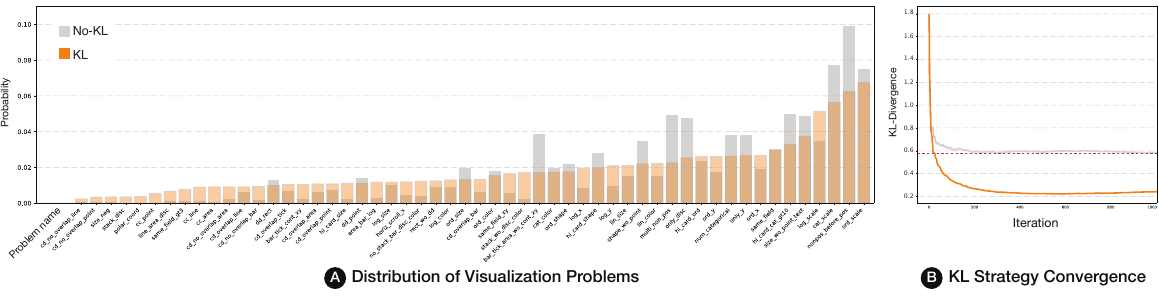}
    
    \caption{\textbf{Effect of KL-divergence filtering during dataset generation}. (\textbf{A}) Distribution of visualization problems across the dataset before (gray) and after (orange) applying KL filtering. The filter substantially reduces skew, leading to a more balanced coverage of problem categories. (\textbf{B}) Evolution of KL-divergence over iterations with and without filtering. The KL strategy consistently converges toward lower divergence, indicating that the resulting dataset more closely approximates a uniform distribution of problems.}
    
    \label{fig:problem_distribution}
\end{figure}

The random sampling of visualization specifications leads to a skewed distribution, dominated by a small set of common problems. To promote balance, we applied a Kullback–Leibler (KL) divergence filter during dataset generation, retaining only specifications that moved the distribution closer to uniformity. Let $p$ denote the empirical distribution of problems and $u$ the uniform distribution. For each candidate specification, we updated counts to obtain a new distribution $p'$ and corresponding divergence $\mathrm{KL}_{\text{new}} = \sum_i p_i \log \tfrac{p_i}{u_i}$. Candidates were accepted if they reduced divergence by at least $\varepsilon = 10^{-3}$, or probabilistically according to Equation~\eqref{eq:acceptance}, with temperature $t=10^{-4}$.

\begin{equation}
\label{eq:acceptance}
\operatorname{P}(\text{accept}) = \exp!\left(\frac{\mathrm{KL} - \mathrm{KL}_{\text{new}}}{t}\right).
\end{equation}

This criterion ensures steady progress toward uniform coverage while allowing occasional acceptance of non-improving candidates to prevent stagnation (see Figure~\ref{fig:problem_distribution}).




\subsection{Evaluation}
\label{subsec:evaluation}
The evaluation was based on structured prompts designed to elicit consistent model behavior. Each prompt included: (i) a role description (e.g., You are an expert in data visualization design using Vega-Lite), (ii) a list of target problems, (iii) the Vega-Lite specification—encoding design elements such as mark type, encodings, channels, variables, and aggregations—together with up to 50 rows of the data table, and (iv) explicit task instructions and required output format (a list of problem names). Complete prompt examples are provided in Appendix~\ref{app:prompts}.

To mitigate sensitivity to prompt formulation~\cite{alexander2024can}, we created five prompt variants that preserved the same structure but differed in wording. For each Vega-Lite specification, one variant was sampled at random per inference. To reduce variance, we repeated each inference five times and averaged results.



Model outputs were evaluated along two dimensions: (i) correctness with respect to ground-truth problems, and (ii) prompt adherence, defined as producing the required format (e.g., ["log\_x", "c\_c\_line"]). Outputs were post-processed with regular expressions to isolate code, parsed into Python objects, and validated against the expected structure. Any parsing failure or off-format response was counted as non-adherent. Adherence was quantified as \(\textit{accuracy} = \frac{N_{\text{adherent}}}{N_{\text{total}}}.\) Generations exceeding a time threshold—typically due to verbose or malformed outputs—were terminated and marked as failures.



\paragraph{Experimental Setup:} We evaluated open-source models using Ollama, including Llama3.18B, Llama3.23B, Gemma34B, Gemma327B, and GPT-OSS~20B. Each model was tested with its default temperature and configuration. Experiments were conducted over approximately four days on two setups: a workstation with an NVIDIA RTX A4500 (20 GB VRAM), 64 GB RAM, and an Intel Core i9-12900 (24 cores), and a Vast AI instance equipped with an NVIDIA A100 SXM4 (40 GB VRAM), 64 GB RAM, and an AMD EPYC 7532 (32 cores).

\section{Results}

\paragraph{Prompt Adherence:}
We evaluated prompt adherence by running $k=5$ inference repetitions per instance, each time sampling one of five prompt variants to account for sensitivity to wording.
Accuracy varied notably across models. Llama3.1 8B (0.65) and Llama3.2 3B (0.87) achieved the lowest adherence, while Gemma3 4B (1.00), Gemma3 27B (1.00), and GPT-OSS~20B (0.98) reached near-perfect scores. These results indicate that Gemma and GPT models reliably produced outputs in the required format, whereas Llama models were more prone to deviations.

High adherence in Gemma and GPT-OSS ensured consistent parsing and reliable correctness evaluation, while the lower adherence of Llama variants introduced risks of parsing errors. Overall, these findings underscore that adherence is a critical factor for the validity of the evaluation pipeline.

\paragraph{ASP vs. Human-Written Problem Definitions:}
We examined whether the formulation of visualization design problems—as Answer Set Programming (ASP) constraints or human-readable descriptions—affects LLM performance. While Draco encodes problems in ASP, these technical formulations are not easily interpretable by language models. To address this, we translated the same set of problems into natural language rules, preserving semantics but aligning with human reasoning.

Experiments were run on the full 2,000-instance dataset using Llama3.2 3B and Gemma3 4B, averaging results over $k=3$ repetitions. With ASP-based prompts, Llama3.2 3B achieved a mean F1-score of 0.073 and Gemma3 4B scored 0.058. Using human-readable formulations, scores improved to 0.082 and 0.145, corresponding to relative gains of 13\% and 150\%, respectively.
These findings indicate that natural language formulations substantially improve LLM performance compared to ASP, underscoring the importance of adapting rule representations to the strengths of LLMs.

\paragraph{Problem Detection:}
We assessed each model’s ability to detect visualization design problems from Vega-Lite specifications using the full dataset of 2,000 instances. For every specification, five prompt variants were randomly sampled across $k=5$ repetitions. F1-scores were computed per problem and averaged, with additional aggregation over broader problem categories (Appendix~\ref{app:problems}) to evaluate generalization across issue types.

Table~\ref{tab:results} reports mean F1-scores and standard deviations for each model and category. GPT-OSS achieved the highest overall performance by far. Gemma 3 27B ranks second overall (0.23 Global Avg.) with strong results on several mark, scale and encoding related items, while Gemma 3 4B outperforms the Llama variants and is comparatively stronger on selected data problems. These results suggest that larger models generally yield more reliable detection, though smaller Gemma models remain competitive in specific domains.

\begin{table}[htbp]
\label{tab:results}
\scriptsize
\centering

\caption{F1-scores (mean) and standard deviations (STD) for open-source LLMs detecting visualization problems and aggregated by categories of problems (\textbf{Avg.}). Average over all problems is showed in \textbf{Global Avg.} for each model. Bold values indicate the best score. Definitions of each problem is available in Appendix~\ref{app:problems}}.

\begin{tabular}{llllllllllll}
\toprule

 & & \multicolumn{2}{c}{Llama 3.1 8B} & \multicolumn{2}{c}{Llama 3.2 3B} & \multicolumn{2}{c}{Gemma3 4B} & \multicolumn{2}{c}{Gemma3 27B} & \multicolumn{2}{c}{GPT-oss 20B} \\
 
\cmidrule(r){3-4} \cmidrule(r){5-6} \cmidrule(r){7-8} \cmidrule(r){9-10} \cmidrule(r){11-12}

& Problem Name & F1 $\uparrow$ & STD & F1 $\uparrow$ & STD & F1 $\uparrow$ & STD & F1 $\uparrow$ & STD & F1 $\uparrow$ & STD \\

\midrule
\multirow{12}{*}{\rotatebox[origin=c]{90}{\makebox[1cm]{encoding}}} 

& size\_negative & 0.05 & 0.00 & 0.05 & 0.01 & 0.04 & 0.00 & 0.35 & 0.01 & \textbf{0.98} & 0.02 \\
& shape\_without\_point & 0.20 & 0.01 & 0.23 & 0.02 & 0.36 & 0.00 & 0.40 & 0.01 & \textbf{0.99} & 0.00 \\
& size\_without\_point\_text & 0.33 & 0.01 & 0.32 & 0.01 & 0.41 & 0.00 & 0.12 & 0.01 & \textbf{1.00} & 0.00 \\
& same\_field\_x\_and\_y & 0.16 & 0.03 & 0.12 & 0.01 & 0.11 & 0.01 & 0.02 & 0.01 & \textbf{0.98} & 0.01 \\
& same\_field & 0.20 & 0.00 & 0.19 & 0.01 & 0.29 & 0.01 & 0.37 & 0.01 & \textbf{0.82} & 0.01 \\
& same\_field\_grt3 & 0.07 & 0.02 & 0.06 & 0.01 & 0.05 & 0.00 & 0.29 & 0.02 & \textbf{0.87} & 0.01 \\
& rect\_without\_d\_d & 0.12 & 0.02 & 0.07 & 0.02 & 0.19 & 0.01 & 0.66 & 0.00 & \textbf{0.90} & 0.01 \\
& number\_categorical & 0.18 & 0.03 & 0.11 & 0.01 & 0.24 & 0.01 & 0.18 & 0.01 & \textbf{0.31} & 0.03 \\
& only\_discrete & 0.14 & 0.03 & 0.09 & 0.01 & 0.20 & 0.01 & 0.01 & 0.00 & \textbf{0.83} & 0.01 \\
& only\_y & 0.13 & 0.02 & 0.04 & 0.01 & 0.20 & 0.00 & 0.02 & 0.00 & \textbf{0.98} & 0.00 \\
& multi\_non\_pos & 0.14 & 0.02 & 0.07 & 0.01 & 0.25 & 0.01 & 0.00 & 0.00 & \textbf{0.93} & 0.01 \\
& non\_pos\_used\_before\_pos & 0.12 & 0.02 & 0.07 & 0.01 & \textbf{0.34} & 0.01 & 0.00 & 0.00 & 0.22 & 0.01 \\

\rowcolor{gray!20} 
& Avg. & 0.15 & 0.02 & 0.12 & 0.01 & 0.22 & 0.01 & 0.20 & 0.01 & \textbf{0.82} & 0.01 \\

\midrule
\multirow{18}{*}{\rotatebox[origin=c]{90}{\makebox[1cm]{mark}}} 

& line\_area\_with\_discrete & 0.08 & 0.01 & 0.06 & 0.01 & 0.05 & 0.00 & 0.49 & 0.01 & \textbf{0.90} & 0.02 \\
& bar\_tick\_continuous\_x\_y & 0.13 & 0.02 & 0.10 & 0.01 & 0.10 & 0.00 & 0.10 & 0.02 & \textbf{0.87} & 0.01 \\
& bar\_tick\_area\_line\_without\_continuous\_x\_y & 0.14 & 0.00 & 0.12 & 0.01 & 0.15 & 0.00 & 0.07 & 0.01 & \textbf{0.80} & 0.01 \\
& area\_bar\_with\_log & 0.13 & 0.01 & 0.10 & 0.01 & 0.21 & 0.00 & 0.51 & 0.01 & \textbf{0.99} & 0.00 \\
& c\_c\_point & 0.10 & 0.03 & 0.03 & 0.01 & 0.07 & 0.01 & 0.39 & 0.01 & \textbf{0.91} & 0.00 \\
& c\_c\_line & 0.14 & 0.04 & 0.05 & 0.03 & 0.08 & 0.00 & 0.50 & 0.00 & \textbf{0.93} & 0.00 \\
& c\_c\_area & 0.17 & 0.05 & 0.05 & 0.02 & 0.12 & 0.01 & 0.36 & 0.01 & \textbf{0.90} & 0.01 \\
& d\_d\_point & 0.01 & 0.01 & 0.03 & 0.02 & 0.07 & 0.01 & 0.45 & 0.01 & \textbf{0.47} & 0.01 \\
& d\_d\_rect & 0.06 & 0.02 & 0.04 & 0.02 & 0.04 & 0.02 & 0.27 & 0.02 & \textbf{0.51} & 0.03 \\
& c\_d\_overlap\_point & 0.12 & 0.02 & 0.08 & 0.01 & \textbf{0.13} & 0.00 & 0.08 & 0.01 & 0.04 & 0.01 \\
& c\_d\_overlap\_bar & 0.02 & 0.02 & 0.06 & 0.03 & \textbf{0.09} & 0.00 & 0.01 & 0.01 & 0.04 & 0.02 \\
& c\_d\_overlap\_line & 0.06 & 0.01 & 0.05 & 0.03 & \textbf{0.11} & 0.01 & 0.03 & 0.01 & 0.10 & 0.01 \\
& c\_d\_overlap\_area & 0.05 & 0.04 & 0.05 & 0.03 & \textbf{0.17} & 0.01 & 0.14 & 0.01 & 0.03 & 0.02 \\
& c\_d\_overlap\_tick & 0.02 & 0.02 & 0.05 & 0.01 & 0.05 & 0.01 & \textbf{0.62} & 0.00 & 0.11 & 0.03 \\
& c\_d\_no\_overlap\_point & 0.06 & 0.03 & 0.03 & 0.01 & 0.03 & 0.01 & 0.00 & 0.00 & \textbf{0.32} & 0.08 \\
& c\_d\_no\_overlap\_bar & 0.01 & 0.02 & 0.02 & 0.02 & 0.03 & 0.01 & 0.00 & 0.00 & \textbf{0.24} & 0.06 \\
& c\_d\_no\_overlap\_line & 0.02 & 0.04 & 0.02 & 0.01 & 0.01 & 0.00 & 0.11 & 0.04 & \textbf{0.34} & 0.03 \\
& c\_d\_no\_overlap\_area & 0.02 & 0.03 & 0.05 & 0.04 & 0.10 & 0.01 & 0.08 & 0.01 & \textbf{0.12} & 0.01 \\

\rowcolor{gray!20} 
& Avg. & 0.07 & 0.02 & 0.05 & 0.02 & 0.09 & 0.01 & 0.23 & 0.01 & \textbf{0.48} & 0.02 \\

\midrule
\multirow{3}{*}{\rotatebox[origin=c]{90}{\makebox[1cm]{stack}}} 

& no\_stack\_with\_bar\_area\_discrete\_color & 0.10 & 0.04 & 0.08 & 0.01 & 0.14 & 0.01 & 0.01 & 0.01 & \textbf{0.85} & 0.01 \\
& stack\_without\_discrete\_color\_or\_detail & 0.18 & 0.02 & 0.10 & 0.02 & 0.16 & 0.00 & 0.23 & 0.01 & \textbf{0.98} & 0.00 \\
& stack\_discrete & 0.05 & 0.02 & 0.03 & 0.01 & 0.04 & 0.00 & 0.03 & 0.02 & \textbf{0.90} & 0.02 \\

\rowcolor{gray!20} 
& Avg. & 0.11 & 0.02 & 0.07 & 0.02 & 0.11 & 0.00 & 0.09 & 0.02 & \textbf{0.91} & 0.01 \\

\midrule
\multirow{15}{*}{\rotatebox[origin=c]{90}{\makebox[1cm]{scale}}} 

& log\_scale & 0.31 & 0.02 & 0.07 & 0.01 & 0.49 & 0.01 & 0.33 & 0.00 & \textbf{1.00} & 0.00 \\
& log\_x & 0.33 & 0.01 & 0.12 & 0.02 & 0.12 & 0.01 & 0.86 & 0.01 & \textbf{0.99} & 0.01 \\
& log\_y & 0.28 & 0.04 & 0.06 & 0.02 & 0.15 & 0.02 & 0.73 & 0.01 & \textbf{0.99} & 0.00 \\
& ordinal\_scale & 0.31 & 0.01 & 0.06 & 0.01 & 0.27 & 0.01 & 0.04 & 0.00 & \textbf{0.90} & 0.00 \\
& ordinal\_x & 0.11 & 0.02 & 0.07 & 0.02 & 0.07 & 0.02 & 0.70 & 0.01 & \textbf{0.86} & 0.01 \\
& ordinal\_y & 0.09 & 0.01 & 0.03 & 0.01 & 0.04 & 0.01 & 0.56 & 0.01 & \textbf{0.83} & 0.01 \\
& categorical\_scale & 0.16 & 0.01 & 0.00 & 0.00 & 0.28 & 0.01 & 0.01 & 0.00 & \textbf{0.49} & 0.04 \\
& categorical\_color & 0.04 & 0.02 & 0.03 & 0.01 & 0.05 & 0.01 & 0.04 & 0.01 & \textbf{0.58} & 0.01 \\
& linear\_color & 0.10 & 0.02 & 0.03 & 0.01 & 0.07 & 0.01 & 0.71 & 0.00 & \textbf{0.91} & 0.01 \\
& linear\_size & 0.08 & 0.02 & 0.04 & 0.02 & 0.03 & 0.01 & 0.74 & 0.01 & \textbf{0.89} & 0.01 \\
& log\_color & 0.38 & 0.06 & 0.11 & 0.02 & 0.15 & 0.03 & 0.89 & 0.00 & \textbf{0.99} & 0.00 \\
& log\_size & 0.23 & 0.04 & 0.09 & 0.02 & 0.09 & 0.02 & 0.90 & 0.01 & \textbf{0.93} & 0.01 \\
& ordinal\_color & 0.04 & 0.02 & 0.05 & 0.01 & 0.04 & 0.02 & 0.70 & 0.00 & \textbf{0.89} & 0.01 \\
& ordinal\_size & 0.02 & 0.01 & 0.04 & 0.01 & 0.03 & 0.01 & 0.83 & 0.00 & \textbf{0.85} & 0.01 \\
& ordinal\_shape & 0.01 & 0.01 & 0.03 & 0.02 & 0.06 & 0.01 & 0.68 & 0.01 & \textbf{0.89} & 0.01 \\

\rowcolor{gray!20} 
& Avg. & 0.17 & 0.02 & 0.05 & 0.01 & 0.13 & 0.02 & 0.58 & 0.01 & \textbf{0.87} & 0.01 \\

\midrule
\multirow{6}{*}{\rotatebox[origin=c]{90}{\makebox[1cm]{data}}} 

& high\_cardinality\_ordinal & 0.19 & 0.01 & 0.11 & 0.01 & \textbf{0.23} & 0.01 & 0.03 & 0.01 & 0.00 & 0.00 \\
& high\_cardinality\_categorical\_grt10 & 0.17 & 0.02 & 0.17 & 0.01 & 0.26 & 0.01 & 0.02 & 0.01 & \textbf{0.51} & 0.03 \\
& high\_cardinality\_shape & 0.09 & 0.02 & 0.10 & 0.03 & 0.19 & 0.01 & 0.22 & 0.02 & \textbf{0.90} & 0.00 \\
& high\_cardinality\_size & 0.06 & 0.04 & 0.04 & 0.02 & \textbf{0.13} & 0.01 & 0.00 & 0.00 & 0.01 & 0.01 \\
& horizontal\_scrolling\_x & 0.06 & 0.03 & 0.06 & 0.02 & \textbf{0.11} & 0.01 & 0.00 & 0.00 & 0.00 & 0.00 \\
& polar\_coordinate & 0.00 & 0.00 & 0.03 & 0.03 & 0.02 & 0.02 & 0.01 & 0.02 & \textbf{0.35} & 0.10 \\

\rowcolor{gray!20} 
& Avg. & 0.09 & 0.02 & 0.08 & 0.02 & 0.16 & 0.01 & 0.05 & 0.01 & \textbf{0.30} & 0.02 \\

\midrule
\rowcolor{gray!10} 
& Global Avg. & 0.12 & 0.02 & 0.07 & 0.02 & 0.14 & 0.01 & 0.23 & 0.01 & \textbf{0.68} & 0.01 \\

\bottomrule
\end{tabular}
\end{table}

\section{Conclusions and Future Work}
\label{sec:conclusions}

In this work, we presented the first systematic evaluation of large language models for detecting visualization design problems directly from Vega-Lite specifications. To enable this study, we introduced a dataset of 2,000 instances covering a broad range of visualization issues, together with standardized prompts and evaluation metrics. Our experiments demonstrate that while recent models such as Gemma 3 and GPT-oss show strong performance and high prompt adherence, smaller Llama variants struggle to consistently follow output instructions, limiting their downstream reliability.

The results highlight two main findings. First, prompt adherence is a critical prerequisite: models that fail to produce outputs in the expected format cannot be reliably evaluated, regardless of their reasoning ability. Second, performance varies across categories of visualization problems, with larger models generally showing stronger generalization but smaller models remaining competitive in certain domains.

Looking forward, several directions are worth exploring. Future work could (i) extend the dataset beyond 2,000 instances to cover more diverse visualization tasks and real-world datasets, (ii) evaluate additional closed-source models and emerging architectures, (iii) investigate fine-tuning or instruction-tuning strategies to improve robustness to prompt variations, and (iv) integrate evaluation pipelines that go beyond F1-score to capture partial correctness and reasoning quality. Finally, connecting these evaluations to downstream applications, such as automated chart auditing tools or visualization recommendation systems, represents an exciting opportunity to bridge research findings with practical impact.

\textbf{Acknowledgments}. This project was supported by Universidad Torcuato Di Tella,
Argentina and Alfred P. Sloan Foundation, Grant G-2024-22665.

\bibliographystyle{plain}
\bibliography{references}

\section{Appendix}

\subsection{Prompts}
\label{app:prompts}

We provide the exact prompt templates used in our experiments. Each template contains placeholders that are replaced at runtime: \texttt{\{vega-spec\}} denotes the Vega-Lite specification under evaluation, and \texttt{\{problems\}} corresponds to the full list of candidate problem names considered in our study. 

\lstset{
  basicstyle=\ttfamily\small,
  breaklines=true,
  frame=single,
  columns=fullflexible
}

\textbf{Prompt \#1}
\begin{lstlisting}[language=]
You are an expert in data visualization design using Vega-Lite.

## Input

### Problems:
{problems}

### Vega-Lite specification:
{vega-spec}

## Task
Analyze the Vega-Lite specification and identify which **exact** problem name from the list above are present.

## Output Requirements
- Output **only** a valid JSON array of strings.
- Each string must be an **exact match** to a problem name from the provided list (excluding the "name :" prefix).
- Do **not** add explanations, reasoning, or any extra text.
- If no problems match, return an empty JSON array: []

### Example:
["problem_A", "problem_B"]
\end{lstlisting}

\textbf{Prompt \#2}
\begin{lstlisting}[language=]
### Problems:
{problems}

You are a Vega-Lite visualization evaluator. Your goal is to read the given Vega-Lite specification and identify any problems from the above list that it exhibits. Focus only on exact matches from the provided names.

### Vega-Lite specification:
{vega-spec}

## Output Requirements
- Output **only** a JSON array of strings.
- Strings must exactly match a problem name from the list (omit "name :").
- No explanations, commentary, or extra formatting.
- If no problems are present, return []

### Example:
["problem_A", "problem_B"]
"""
\end{lstlisting}

\newpage

\textbf{Prompt \#3}
\begin{lstlisting}[language=]
### Problems:
{problems}

Analyze the following Vega-Lite specification carefully. Report which of the problem names listed above are present in it. Only use exact matches.

### Vega-Lite specification:
{vega-spec}

## Output Requirements
- Return **only** a JSON array of strings.
- Each string must match a problem name exactly (exclude "name :").
- Do not include explanations, notes, or any additional text.
- Return [] if there are no matches.

### Example:
["problem_A", "problem_B"]
"""
\end{lstlisting}

\textbf{Prompt \#4}
\begin{lstlisting}[language=]
### Problems:
{problems}

You are tasked with checking the Vega-Lite specification below. Identify all problems from the above list that appear in it. Ensure each match is exact.

### Vega-Lite specification:
{vega-spec}

## Output Requirements
- Provide **only** a JSON array of strings.
- Each string must be an exact problem name (without "name :").
- No extra text, reasoning, or commentary.
- If no problems are found, return []

### Example:
["problem_A", "problem_B"]
"""
\end{lstlisting}

\textbf{Prompt \#5  }
\begin{lstlisting}[language=]
### Problems:
{problems}

Review the Vega-Lite specification and determine which problem names from the list are present. Only include exact matches in your output.

### Vega-Lite specification:
{vega-spec}

## Output Requirements
- Output **only** a JSON array of strings.
- Strings must exactly match the problem names (ignore the "name :" prefix).
- Do not provide explanations, notes, or any additional text.
- If none match, return []

### Example:
["problem_A", "problem_B"]
"""
\end{lstlisting}

\subsection{Problems}

We provide here the full list of visualization problems used in our evaluation written in natural language. Each entry includes the problem \textbf{name} and \textbf{description}. The same text was provided to LLMs during evaluation.

\begin{lstlisting}[language=]
name: size_negative
A violation occurs if a channel is encoded with size and the corresponding field contains both negative and positive values. The size encoding implies positive magnitude, so it should not be used when the data includes negative values.

name: line_area_with_discrete
A violation occurs when a line or area chart is used and both the x and y channels are encoded with discrete scales. Line and area marks are intended for continuous data and do not function correctly with fully discrete axes.

name: bar_tick_continuous_x_y
A violation occurs if a bar or tick chart is used and both the x and y channels are continuous. These mark types are designed to compare discrete categories and are not suitable for continuous continuous configurations.

name: shape_without_point
A violation occurs when the shape channel is used but the mark type is not point. The shape encoding is only meaningful when applied to point marks.

name: size_without_point_text
A violation occurs if the size channel is used with a mark type that is neither point nor text. Only point and text marks properly support the size encoding.

name: area_bar_with_log
A violation occurs when a bar or area chart uses a logarithmic scale on either the x or y axis. Using log scales with these mark types can produce misleading visuals and should be avoided.

name: rect_without_d_d
A violation occurs if a rect mark is used and either the x or y channel is continuous. Rect marks require both axes to be discrete to represent a meaningful tiled layout.

name: same_field_x_and_y
A violation occurs when the same field is assigned to both the x and y channels of a single mark. This redundancy creates a chart that is either meaningless or visually confusing.

name: bar_tick_area_line_without_continuous_x_y
A violation occurs when a chart uses a bar, tick, area, or line mark but neither the x nor y channel is continuous. These marks depend on at least one continuous axis to effectively display measurements or trends.

name: no_stack_with_bar_area_discrete_color
A violation occurs when a bar or area chart uses a discrete or binned color channel but does not use stacking. Stacking is required to accurately represent grouped values in this context.

name: stack_without_discrete_color_or_detail
A violation occurs when stacking is enabled on a mark, but neither a discrete/binned color channel nor a detail channel is used. Stacking requires at least one of these to define how data should be grouped.

name: stack_discrete
A violation occurs when stacking is applied to a channel that is discrete or binned. Stacking must only be applied to continuous channels to ensure correct rendering of data aggregation.

name: same_field
Triggers when the same field is used exactly twice as an encoding for the same mark. This indicates a preference to avoid duplicating the same data field in multiple channels for a single mark.

name: same_field_grt3
Triggers when the same field is used three or more times as an encoding for the same mark. This indicates a stronger penalty for repeatedly using the same field excessively.

name: number_categorical
Triggers when a field of type `number` is encoded with a categorical scale type. This reflects a preference against treating numeric data as categorical.

name: only_discrete
Triggers when a mark has no continuous encodings all its channels are discrete or binned.

name: multi_non_pos
Triggers when a single mark uses more than one non-positional channel (e.g., color, size, shape).

name: non_pos_used_before_pos
Triggers when a non-positional channel is used in a mark but neither `x` nor `y` positional channels are present.

name: only_y
Triggers when a mark has an encoding for `y` but no encoding for `x`.

name: high_cardinality_ordinal
Triggers when a field encoded with an ordinal scale has cardinality greater than 30.

name: high_cardinality_categorical_grt10
Triggers when a field encoded with a categorical scale has cardinality greater than 10.

name: high_cardinality_shape
Triggers when the shape channel is encoded with a field having cardinality greater than 8.

name: high_cardinality_size
Triggers when the size channel is present, and the `x` or `y` positional encoding is continuous and has cardinality greater than 100.

name: horizontal_scrolling_x
Triggers when the x-channel is discrete or binned and has cardinality greater than 50.

name: log_scale
Triggers when an encoding uses a log scale type.

name: ordinal_scale
Triggers when an encoding uses an ordinal scale type.

name: categorical_scale
Triggers when an encoding uses a categorical scale type.

name: c_c_line
Triggers when both x and y are continuous and the mark type is `line`.

name: c_c_area
Triggers when both x and y are continuous and the mark type is `area`.

name: c_d_overlap_point
Triggers when the x/y relationship is continuous by discrete, overlap is detected, and the mark type is `point`.

name: c_d_overlap_bar
Triggers when the x/y relationship is continuous by discrete, overlap is detected, and the mark type is `bar`.

name: c_d_overlap_line
Triggers when the x/y relationship is continuous by discrete, overlap is detected, and the mark type is `line`.

name: c_d_overlap_area
Triggers when the x/y relationship is continuous by discrete, overlap is detected, and the mark type is `area`.

name: c_d_no_overlap_point
Triggers when the x/y relationship is continuous by discrete, no overlap is detected, and the mark type is `point`.

name: c_d_no_overlap_line
Triggers when the x/y relationship is continuous by discrete, no overlap is detected, and the mark type is `line`.

name: c_d_no_overlap_area
Triggers when the x/y relationship is continuous by discrete, no overlap is detected, and the mark type is `area`.

name: linear_color
Triggers when the color channel is used with a linear scale type.

name: linear_size
Triggers when the size channel is used with a linear scale type.

name: log_color
Triggers when the color channel is used with a log scale type.

name: log_size
Triggers when the size channel is used with a log scale type.

name: ordinal_x
Triggers when the x-channel is used with an ordinal scale type.

name: ordinal_color
Triggers when the color channel is used with an ordinal scale type.

name: ordinal_size
Triggers when the size channel is used with an ordinal scale type.

name: ordinal_shape
Triggers when the shape channel is used with an ordinal scale type.

name: categorical_color
Triggers when the color channel is used with a categorical scale type.

name: polar_coordinate
Triggers when the view coordinates are set to `polar`.

name: c_c_point
Triggers when both x and y are continuous and the mark type is `point`.

name: c_d_overlap_tick
Triggers when the x/y relationship is continuous by discrete, overlap is detected, and the mark type is `tick`.

name: c_d_no_overlap_bar
Triggers when the x/y relationship is continuous by discrete, no overlap is detected, and the mark type is `bar`.

name: d_d_point
Triggers when both x and y are discrete and the mark type is `point`.

name: d_d_rect
Triggers when both x and y are discrete and the mark type is `rect`.

name: log_x
Triggers when the x-channel uses a log scale type.

name: log_y
Triggers when the y-channel uses a log scale type.

name: ordinal_y
Triggers when the y-channel uses an ordinal scale type.
\end{lstlisting}

\label{app:problems}

\end{document}